\pdfoutput=1

\documentclass[11pt]{article}

\usepackage[]{emnlp2021}

\usepackage{times}
\usepackage{latexsym}

\usepackage[T1]{fontenc}

\usepackage[utf8]{inputenc}

\usepackage{microtype}
\usepackage{graphicx}
\usepackage{amsmath}
\usepackage{amssymb}
\title{A Grounded Well-being Conversational Agent\\ with Multiple Interaction Modes: Preliminary Results}

\author{Xinxin Yan, \and Ndapa Nakashole \\
        Computer Science and Engineering \\
        University of California, San Diego \\
        La Jolla, CA 92093 \\
        \texttt{x3yan@ucsd.edu, nnakashole@eng.ucsd.edu}}

\begin{document}
\maketitle
\begin{abstract}
Technologies for   enhancing  well-being,  healthcare vigilance and monitoring  are on the rise.
However, despite  patient interest, such technologies suffer from low adoption. One hypothesis for this limited adoption is loss of  human interaction that is central to  doctor-patient encounters. In this paper we seek to address this limitation via a conversational agent that adopts one aspect of in-person doctor-patient interactions: A human avatar to facilitate medical grounded question answering. This is akin to the in-person scenario where the doctor may point to the human body or the patient may point to their own body to express their conditions.  Additionally, our agent has multiple interaction modes, that may give more options for the patient to use the agent, not just for medical question answering, but also to engage in  conversations about general topics and  current events. Both the avatar, and the multiple interaction modes could help improve adherence.

We  present a high level overview of the design of our agent,  Marie Bot Wellbeing. We also report implementation details of our early prototype , and present preliminary results.
\end{abstract}

\section{Introduction}



NLP is in a position to bring-forth scalable, cost-effective solutions for promoting  well-being. Such solutions can serve many segments of the  population such as people living in medically under-served communities with limited access to clinicians, and people  with limited mobility.  These solutions can also serve those interested in  self-monitoring~\cite{torous2014smartphone} their own health. There is  evidence that these technologies can be effective \cite{mayo2007internet,fitzpatrick2017delivering}. However, despite  interest, such technologies suffer from low adoption\cite{donkin2013rethinking}. One hypothesis for this limited adoption is the loss of human interaction which is central to  doctor-patient encounters\cite{fitzpatrick2017delivering}. In this paper we seek to address this limitation via a conversational agent that emulates one aspect of in-person doctor-patient interactions: a human avatar to facilitate grounded question answering. This is akin to the in-person scenario where the doctor may point to the human body or the patient may point to their own body to express their conditions.  Additionally, our agent has multiple interaction modes, that may give more options for the patient to use the agent, not just for medical question answering, but also to engage in  conversations about general topics and  current  events.  Both the avatar, and the multiple interaction modes could help improve adherence.


The human body is complex and information about how it functions fill entire books. Yet it is important for individuals to know about conditions that can affect the human body, in order to practice continued monitoring and prevention to keep severe medical situations at bay. To this end, our well-being agent {includes} a medical question answering interaction mode (\textbf{MedicalQABot}).
 For   mental health,  social isolation and loneliness can have adverse health consequences such as  anxiety, depression, and suicide. Our well-being agent includes a social interaction mode (\textbf{SocialBot}), wherein the agent can be an approximation of human  a companion.
The MedicalQABot is less conversational but accomplishes the task of answering questions. The SocialBot seeks to be conversational while providing some information. And, there is a third interaction mode, the \textbf{Chatbot}, which in our work is used as a last-resort mode, it is conversational but does not provide much information of substance.

To test the ideas of our proposed agent, we are developing a grounded well-being conversational agent, called  ``Marie Bot Wellbeing".
This paper  presents   a sketch of the high level  design of our Marie system, and some preliminary results.

\begin{figure*}[t]
	\centering
			\vspace{-1.2cm} 
	\includegraphics[width=0.8\linewidth]{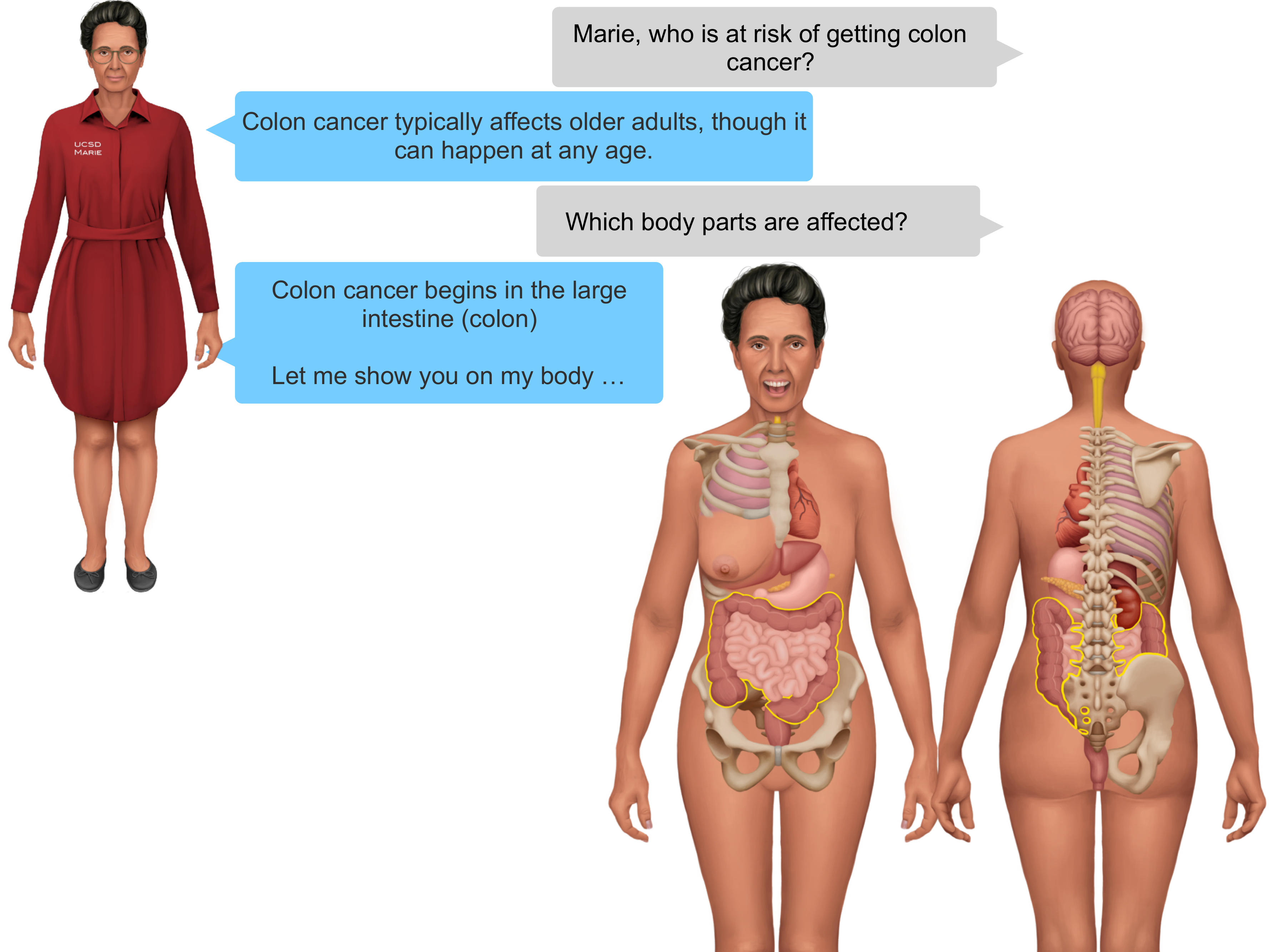}
	\caption{An illustration of the MedicalQA interaction mode. Here the agent's answer is grounded on our human avatar. The affected body part, the large intestine, is highlighted on the avatar.}
	\label{fig:examplemedqa}
\end{figure*}

An important consideration when developing technology for healthcare is that there is \textit{low tolerance for errors.}  Erroneous information   can have severe negative consequences. We design the medicalQABot, and the SocialBot with this  consideration in mind.
Our  design philosophy  consists of the following tenets:
\begin{enumerate}
\item \textbf{Reputable answers:} Only provide  answers to questions for which we have  answers from reputable sources, instead of considering information from every corner of the Web.
\item \textbf{Calibrated confidence scores:} 
Even though the answers come from reputable sources, there are various decisions that are involved that the model must make including which specific answer to retrieve for a given question.
For these  predictions by our models, we must know what we do not know, and provide only information about which the model is fairly certain.

\item \textbf{Visualize}: Whenever an answer can be visualized to some degree, we should provide a visualization to accompany the text answer to help clarify, and reduce misunderstanding. 

\item \textbf{Graceful failure}: when one of the interaction modes fails, another interaction mode can take over.
\end{enumerate}

\begin{figure*}[t]
	\centering
			\vspace{-1.2cm} 
	\includegraphics[width=0.8\linewidth]{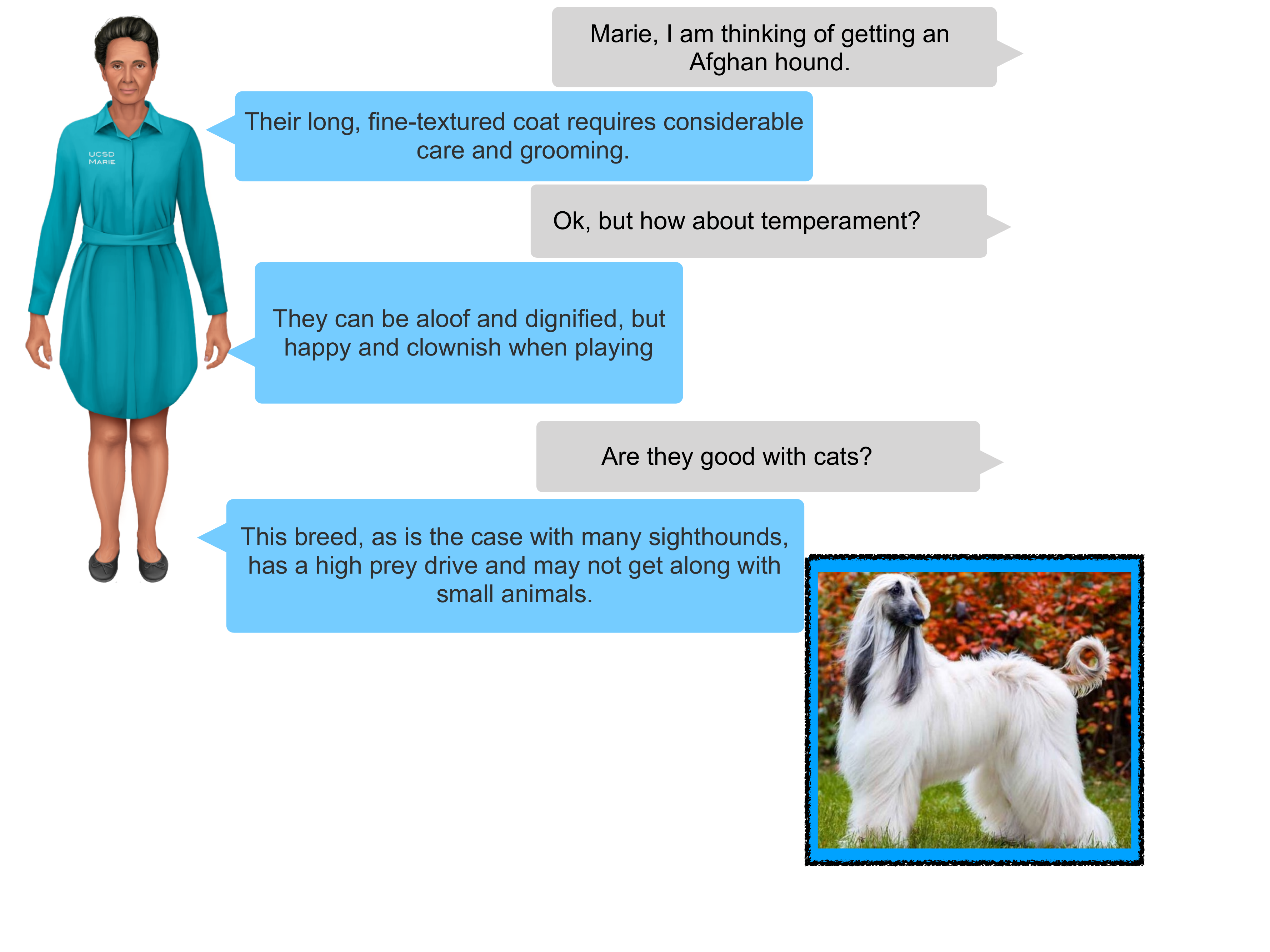}
			\vspace{-1.2cm} 
	\caption{An illustration of the SocialBot interaction mode}
	\label{fig:examplesocialqa}

\end{figure*}
%

\paragraph{Organization}
In what follows, we discuss how the above tenets are manifested in our agent.

The rest of the paper is organized as follows: We begin with a high-level overview of the design of the different parts of the agent (Sections~\ref{sec:modes} to  \ref{sec:modessocial}); We next discuss the current prototype implementation and  preliminary results (Section~\ref{sec:implementation}); We next present related work (Section~\ref{sec:relatedwork}); and close with a discussion  (Section~\ref{sec:discussion}) and concluding remarks (Section~\ref{sec:conclude}).

\section{Interaction Modes and Dialog Management}\label{sec:modes}

 In navigating between the different interaction modes, we design our system as follows.
Based on the user utterance, we automatically predict using a binary classifier  to switch between different interaction modes ( MedicalQABot vs SocialBot).
Suppose that the classifier predicts that the utterance is a question asking for medical information on a topic, and suppose our medicalQA determines that we have no information on that topic, our goal is to then  let the SocialBot take over if it has information on that topic and can meaningfully hold a conversation about it. For the SocialBot, when   missing the necessary information, our goal is to have it fall back to Chatbot mode.
%

\section{MedicalQABot Mode}
\subsection{Knowledge vault of QA pairs}  
 Some aspects of the human body are well-understood, many diseases and medical conditions have been studied for many years. Thus a lot of medical questions have already been asked, and their answers are known. Thus one approach to medicalQA is a retrieval-based one which consists of two steps: First, we collect and create a knowledge vault of frequently asked questions and their curated answers from reputable sources.
 
 
 Second, given a user question, we must match it to one of the questions in the QA knowledge vault.
 However, when  people pose their questions, they are not aware of the exact words used in the questions of the knowledge vault. We must therefore match user questions to the correct question in the knowledge vault. A simple approach is keyword search. However, this misses a lot of compositional effects. One other way is to treat this as a problem of entailment. Where given a user question, we can find,  in the knowledge vault, the questions that entail the user question. 
 
 This mode can  benefit from well-studied aspects of   knowledge harvesting and representation~\cite{DBLP:conf/webdb/NakasholeTW10,theobald2010urdf,DBLP:conf/vlds/NakasholeSST12, nakashole2012automatic,kumar2017discovering}.

\subsection{Grounding to Human Anatomy Avatar}
We develop a human avatar to help users better  understand medical  information. And also to help them to more precisely specify their questions. The avatar is meant to be used in two ways.  The human avatar was illustrated by a medical illustrator we hired from Upwork.com. 

\textbf{Bot $\rightarrow$ Patient:}  When an answer contains body parts, relevant body parts are highlighted on the avatar.
		"this medical condition affects the following body parts ". An illustration of this direction is shown in Figure \ref{fig:examplemedqa}.
		
\textbf{Patient $\rightarrow$ Bot}: When the user describes their condition, they can point by clicking. "I am not feeling well here". 
		

\section{SocialBot Mode}\label{sec:modessocial}
For the SocialBot, we propose to create a  knowledge vault of topics that will enable  the bot to have engaging conversations  with humans on  topics of interest including current events.
For example, the bot can say ``Sure, we can talk about  German beer" or. "I see you want  to talk about Afghan hounds"".
The topics will be mined from Wikipedia, news sources, and social media including Reddit.
For the SocialBot, we wish  to model the principles of  a good conversation: having something interesting to say, and  showing interest in what the conversation partner says   \cite{MariOstendorf}





\section{Prototype Implementation \& Preliminary Experiments}\label{sec:implementation}
Having discussed the high-level  design goals, in the following sections  we present specifics of our initial prototype.
Our prototype's language understanding  capabilities are limited. They can be thought of as placeholders that allowed us to quickly develop a prototype. These simple capabilities will be replaced as we develop more  advanced language  processing  methods for our  system.

\subsection{Data}
We describe the data used in our current prototype.

\paragraph{Medline Data}
We collected Medline data \footnote{https://medlineplus.gov/xml.html}, containing 1031 high level medical topics.  We extracted the summaries and split the text into questions and answers. We generated several data files from this dataset: question-topic pair data, answer-topic pair data and question-answer pair data.
The data size and split  information is presented  in Table~\ref{tab:topic_data}.
We will describe their usage in detail in the following sections

\paragraph{Medical Dialogue Data}
We use  the MedDialog dataset\cite{zeng2020meddialog} which has 0.26 million dialogues between patients and doctors. The raw dialogues were obtained from healthcaremagic.com and icliniq.com. 

We also use the MedQuAD (Medical Question Answering Dataset) dataset \cite{BenAbacha-BMC-2019} which contains 47457 medical question-answer pairs created from 12 NIH\footnote{https://www.nih.gov/} websites. 

\paragraph{News Category Dataset}
We also use the News category dataset from Kaggle\footnote{https://www.kaggle.com/rmisra/news-category-dataset}. It contains 41 topics. 
We use the data in 39 topics, without "Healthy Living" and "Wellness", which might be related to the medical domain. We extract the short description from the dataset.

\begin{figure*}[t]
    \centering
     \vspace{-0.8cm}
    \includegraphics[scale=0.4]{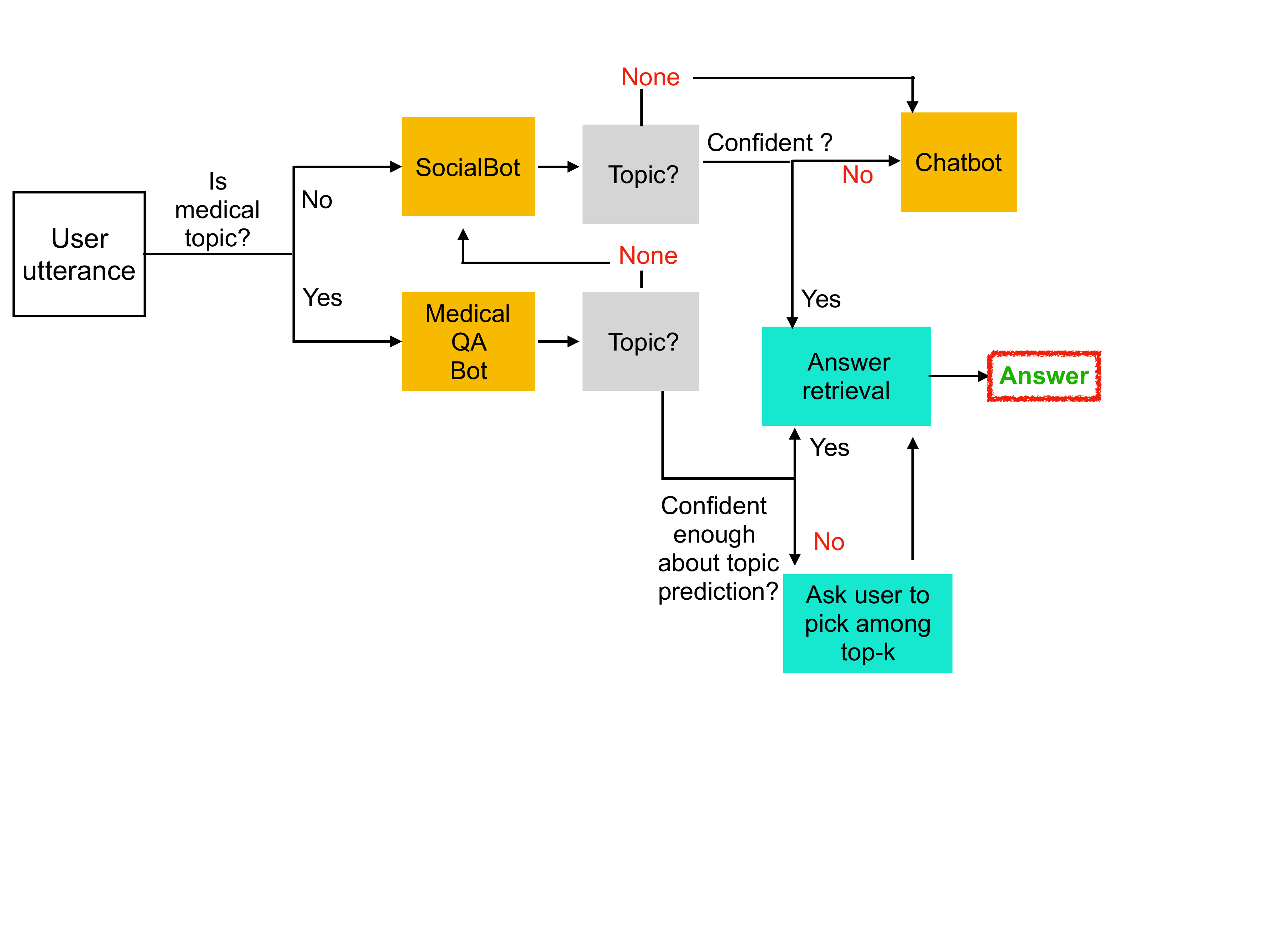}
       \vspace{-3.2cm}
    \caption{Our proposed pipeline.  Section~\ref{sec:implementation} has  more details on the implementation of our current prototype.}
 
    \label{fig:chat_agent_structure}
\end{figure*}

\paragraph{Reddit Data}
We collected questions and comments from 30 subreddits. We treat  each subreddit as one topic. The number of questions for each topic is shown in Table \ref{tab:subreddit-question-num}. This Reddit data is  to be used for our SocialBot.

\subsection{System Overview}
As shown in Figure \ref{fig:chat_agent_structure}, our system makes a number of decisions upon receiving a user utterance. First, the system predicts if the utterance should be handled by the MedicalQABot or by the SocialBot.

If  the MedicalQABot is predicted to handle the utterance, then an additional decision is made. This decision predicts which Medical topic the utterance is about. If we are not certain, the system puts the user in the loop, by asking them to confirm the topic. If the user says the top predicted topic is not the correct one, we present them with the next topic in the order, and ask them again, up to 4 times.

If the SocialBot is predicted to handle the utterance, the goal is to have the system decide between various general topics and current events for which the system has collected information. If the topic is outside of the scope of what the SocialBot knows, the system resorts to a ChatBot, that may just give generic responses, and engage in chitchat dialogue.


\subsection{Mode Prediction  Classifier}
We train this classifier to determine whether the user's input is related to the medical domain. We use the output from BERT encoder as the input to a linear classification layer  trained with a cross-entropy loss function.

We choose the positive examples from MedQuAD Dataset, and negative examples from News Category Dataset. The training data information is shown in Table \ref{tab:med_nonmed_data}. And the evaluation results are shown in Table \ref{tab:med_nonmed_result}. This performance is potentially better than in real-life settings, because the medical (medline) vs non-medical (Kaggle news) data is cleanly separated. In reality, a user utterance might be "I am not happy, I have a headache" they may not want to get medical advise, but simply  to  just chat a bit to distract them from the headache.

\begin{table}[t]
    \centering
    \begin{tabular}{|c|c|}
    \hline
      Train & 286370   \\
         \hline
        Valid & 35796 \\
       \hline
       Test & 35797 \\
       \hline
    \end{tabular}
    \caption{Interaction Mode Prediction   Data }
    \label{tab:med_nonmed_data}
\end{table}

\begin{table}[t]
    \centering
    \begin{tabular}{|c|c|}
    \hline
       
       Valid accuracy & 0.9970 \\
       \hline
       Test accuracy & 0.9972 \\
       \hline
    \end{tabular}
    \caption{Interaction Mode Prediction     Evaluation Results}
    \label{tab:med_nonmed_result}
\end{table}

\subsection{MedicalQA  Implementation }
\paragraph{Medical Topic Classifier}
If the user utterance is routed to the  MedicalQABot,  the  MedicalQABot first  predicts the medical category of the user's input.
We use Medline Data, which contains 1031 topics, to train this classifier. The dataset information is shown in Table~\ref{tab:topic_data}. 
The evaluation results of our medical topic classifier is shown in Table \ref{tab:topic_result}.
\begin{table}[t]
    \centering
    \begin{tabular}{|c|c|}
    \hline
      Train & 12082   \\
         \hline
        Valid & 3021 \\
       \hline
       Test & 615 \\
       \hline
    \end{tabular}
    \caption{Medical Topic Classifier Training Data Information}
    \label{tab:topic_data}
\end{table}

\begin{table}[t]
    \centering
    \begin{tabular}{|c|c|}
    \hline
         Train accuracy & 0.8812 \\
  \hline
       Test accuracy & 0.8358 \\
       \hline
    \end{tabular}
    \caption{Medical Topic Classifier Evaluation Results}
    \label{tab:topic_result}
\end{table}
\paragraph{Topic Posterior Calibration}
As shown in Figure \ref{fig:chat_agent_structure}, we ask a topic confirmation question after the topic classifier, which is used to let the user confirm the correctness of the output from Topic classifier. But we do not always need the confirmation. We set a threshold for the confidence score of the classifier. If the confidence score is higher than the threshold, meaning that our classifier is confident enough in the output, we will skip the confirmation question and retrieve the answer directly.

To make the classifier confidence scores more reliable, we use  posterior calibration to  encourage the confidence level to correspond to the probability that the classifier is correct \cite{calibration,temp-calibration}. The method learns a  parameter, called temperature or $T$.
Temperature is introduced to the output logits of the model as follows:
\begin{equation}
    pred = arg_i max \frac{exp(z_i/T)}{\Sigma_j exp(z_j/T)}
\end{equation}
$\{z_i\}$ is the logits of the model and $T$ is the temperature that needs to be optimized. $T$ is optimized on a validation set to maximize the log-likelihood. 


\paragraph{MedicalQA Retriever}
After we determine the topic of the user's input, we can retrieve the answer from the Medline Dataset. We split the paragraphs in Medline data into single sentences and label them with the topics they belong to. We train the retriever using the augmented Medline data. We split the dataset into train, validation and test set using the ratio 8:1:1. The current retriever is based on BERT NextSentencePrediction model. We use the score from the model to determine the rank of each answer, and concatenate top 3 as the response of the agent.
The evaluation result is shown in Table \ref{tab:retrieve_result}. 
\begin{table}[t]
    \centering
    \begin{tabular}{|c|c|}
    \hline
      Precision & 0.7585 \\
         \hline
        Recall & 0.7621 \\
       \hline
       F-1 score & 0.7603 \\
       \hline
       Accuracy & 0.7597 \\
       \hline
    \end{tabular}
    \caption{MedicalQA Retriever Evaluation Results}
    \label{tab:retrieve_result}
\end{table}

\begin{figure*}[t]
    \centering
    \includegraphics[width=1.0\linewidth]{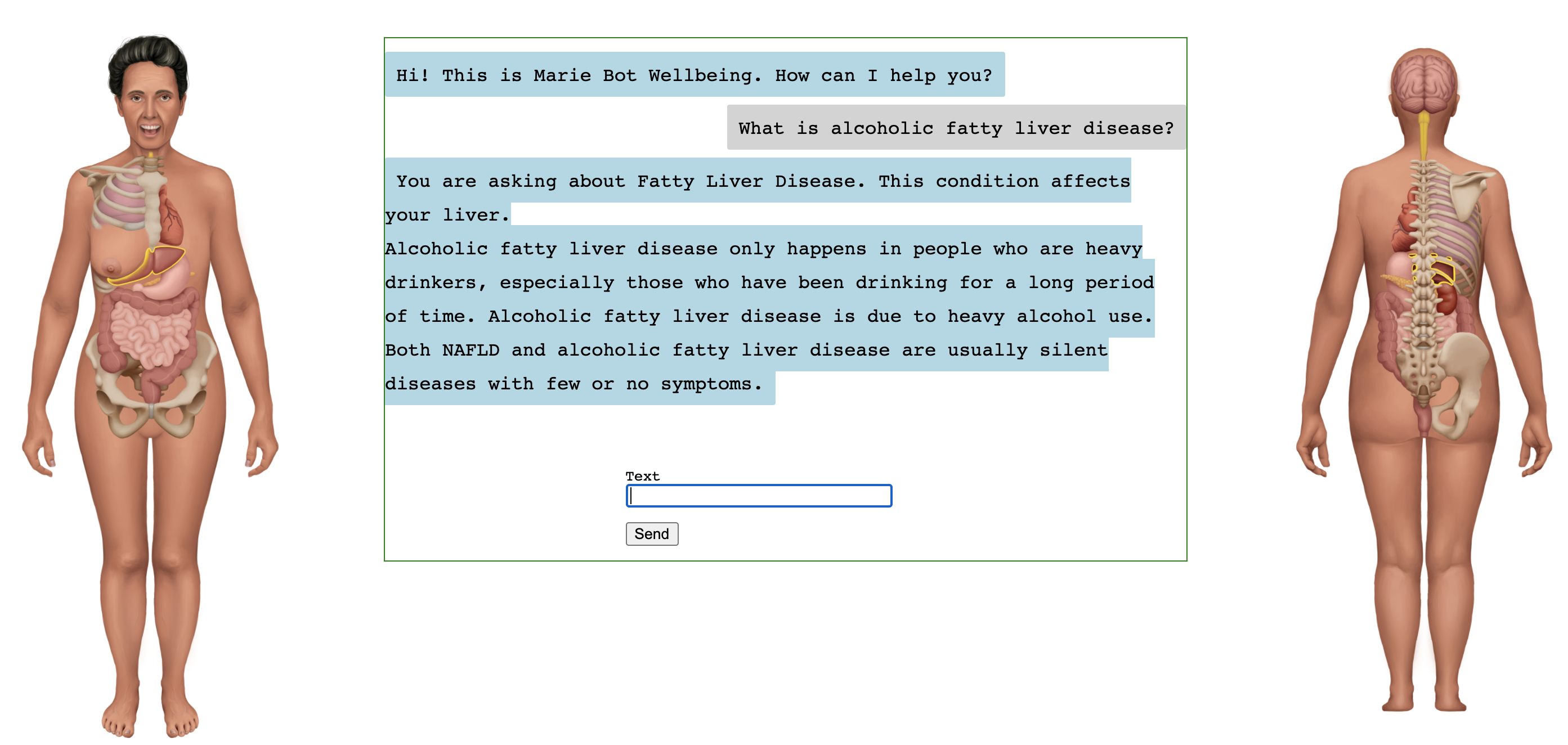}
    \vspace{-0.4cm}
    \caption{Human avatar visual answer example from our prototype: The affected body part, the liver, is highlighted on the avatar.}
    \label{fig:visual_answer}
\end{figure*}

\begin{figure*}[t]
    \centering
    \includegraphics[width=1.0\linewidth]{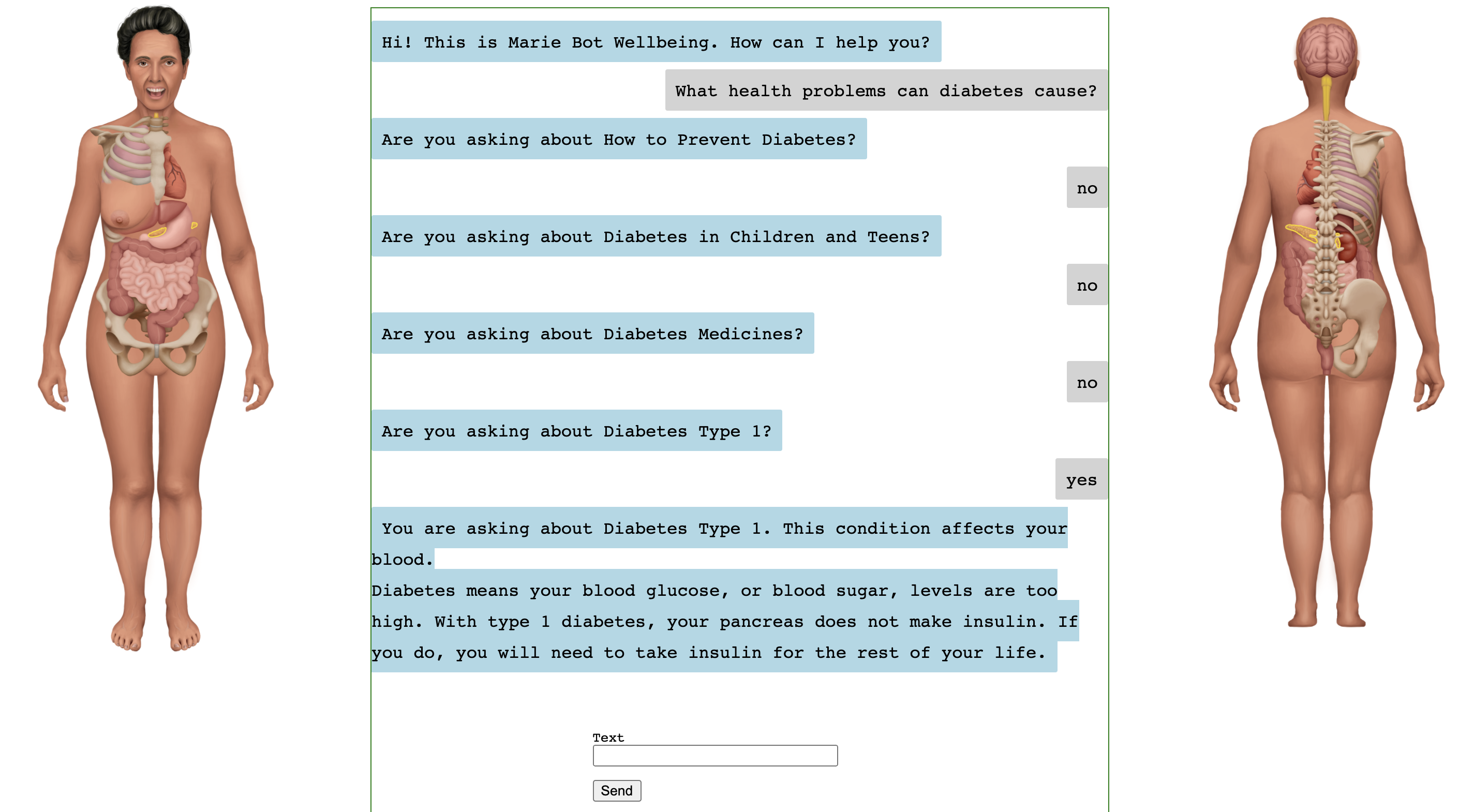}
    \vspace{-0.4cm}
    \caption{Human Avatar Visual Answer Example From our Prototype: Diabetes/Blood Sugar}
    \label{fig:visual_answer2}
\end{figure*}

\subsection{MedicalQA Grounding with Human Avatar}
Our initial version for the human avatar  contains 49 key body parts for front and 33 key body parts for the back. The front and back body part keywords are shown in Table \ref{tab:front_bodyparts} and \ref{tab:back_bodyparts}. As future work, our goal is a  more complete avatar with a comprehensive list of body parts. 

Example grounded answers in our prototype system are shown in Figures ~\ref{fig:visual_answer} and ~\ref{fig:visual_answer2} .


\subsection{SocialBot Implementation}
For our SocialBot, we currently have collected data  from Reddit where each subreddits corresponds to a topic as shown in Table\ref{tab:subreddit-question-num}.
The topic classifier, posterior calibrator, and answer retriever are the same as in the MedicalQABot.

\subsection{ChatBot Implementation}
What is implemented is the last resort ChatBot, for which we have two versions: one  is derived from a base language model, and another derived from a fine-tuned language model. 

\paragraph{Language Models}
We use a large scale pre-trained language model, OpenAI GPT, as our base language model. We use the idea of transfer learning, which starts from a language model pre-trained on a large corpus, and then fine-tuned on end task. This idea was inspired by the huggingface convai project \cite{convai}.



\textit{Fine-tuning on Medical Dialogue Dataset:} We use the Medical Dialogue Data\cite{zeng2020meddialog}  to fine-tune the pre-trained language model. 
We use the questions as chat history and answers as current reply. The training set contains the portion   from healthcaremagic and the test set the portion from icliniq

The evaluation results of our  language model ChatBot are shown in Table \ref{tab:base_LM_result}. 
\begin{table}[t]
    \centering
    \begin{tabular}{|c|c|c|}
    \hline
         & NLL & PPL \\
         \hline
       pre-trained model & 5.4277 & 227.6291\\
       \hline
       fine-tuned model & 3.2750 & 26.4423\\
       \hline
    \end{tabular}
    \caption{Language Model Evaluation. Negative log likelihood (NLL) and Perplexity (PPL)}
    \label{tab:base_LM_result}
\end{table}

\section{Related Work}\label{sec:relatedwork}
\paragraph{Medical Conversational Agents}
Academic~\cite{johnson2020voice} and industry NLP research continues to push the frontiers of conversational agents, for example Meena from Google trained on a large collection of raw text~\cite{open-domain-chatbot}. In that work, it was found that end-to-end neural network with sufficiently low perplexity can surpass the sensibleness and specificity of existing chatbots that rely on complex, handcrafted frameworks. 
Medical dialogue has also been pursued from various angles for automatic diagnosis~\cite{wei2018task,xu2019end}.  

\paragraph{Grounding to Human Avatar}
IBM Research developed a human avatar for patient-doctor interactions \cite{IBM-human-avatar} with a focus on visualizing electronic medical records. By clicking on a  particular body part on the avatar, the doctor can trigger the search of medical records and retrieve relevant information. Their focus on electronic medical records is different from our grounded medical question answering focus.

Another work \cite{human-avatar-close-gap}  analyzed whether and how  the avatars help close the doctor-patient communication gap. This study showed that poor communication between doctors and patients often leads patients to not follow their prescribed treatments regimens. Their thesis is that avatar system can help patients better understanding the doctor's diagnosis. They put medical data, FDA data and user-generated content into a single site that let people search this integrated content by clicking on a virtual body. 

\section{Discussion}\label{sec:discussion}

\subsection{Technical Challenges}

\paragraph{Quality and Quantity of Data}
In order for users to find the agent useful, and for the agent to really have a positive impact,  we must provide answers to more questions. We need to extract more questions from a diverse set of  reputable  sources, while improving coverage.


\paragraph{Comprehensive Visualizations}
For the visualization, and human avatar grounding to be useful,  a more comprehensive avatar is required, with all the  parts that make up the human body. Medical ontologies such as the SNOMED CT part of Unified Medical Language System (UMLS)\footnote{https://www.nlm.nih.gov/research/umls/index.html} contain a comprehensive list of the human body structures, which we can exploit and provide to a medical illustrator.

\subsection{Ethical Considerations}

\paragraph{Privacy}
When we deploy our system, we will respect user privacy, by  not asking for identifiers. Additionally, we will store our data  anonymously. Any real-world  data will only accessible to researchers directly involved with our study.


\paragraph{False Information}
False or erroneous information   in our data sources could lead our agent to present  answers with potentially dire consequences. Our approach of only answering medical questions for which we have high quality, human curated answers seeks to address this concern.

\paragraph{System Capabilities Transparency}
Following  prior work on automated  health systems, our goal is to be clear and transparent about system capabilities~\cite{kretzschmar2019can}.




\section{Conclusion}
\label{sec:conclude}
We have presented a high level overview of the design philosophy of Marie Bot Wellbeing,  a grounded,  multi-interaction mode well-being conversational agent. The agent is designed to mitigate the limited adoption  that plagues  agents for  healthcare despite patient interest.   We  reported  details of our prototype implementation, and preliminary results. 

There is much more to be done to fully realize Marie, which is part of our ongoing work.


\bibliography{anthology,custom}
\bibliographystyle{acl_natbib}

\appendix

\label{sec:appendix}
\section{Appendix}
\begin{table}[t]
    \centering
    \begin{tabular}{|c|c|}
    \hline
      SubReddit & Question Num   \\
         \hline
        AskPhotography & 996 \\
        \hline
        NoStupidQuestions & 912 \\
        \hline
        AskHistorians & 985 \\
        \hline
        askscience & 998 \\
        \hline
        AskWomen & 525 \\
        \hline
        AskReddit & 925 \\
        \hline
        AskUK & 781 \\
        \hline
        AskMen & 200 \\
        \hline
        AskCulinary & 998 \\
        \hline
        AskEconomics & 560 \\
        \hline
        AskAnAmerican & 850 \\
        \hline
        AskALiberal & 830 \\
        \hline
        askaconservative & 775 \\
        \hline
        AskElectronics & 842 \\
        \hline
        Ask\_Politics & 999 \\
        \hline
        AskEngineers & 912 \\
        \hline
        askmath & 999 \\
        \hline
        AskScienceFiction & 652 \\
        \hline
        AskNYC & 994 \\
        \hline
        AskTrumpSupporters & 357 \\
        \hline
        AskDocs & 684 \\
        \hline
        AskAcademia & 987 \\
        \hline
        askcarsales & 995 \\
        \hline
        askphilosophy & 981 \\
        \hline
        AskSocialScience & 487 \\
        \hline
        AskEurope & 844 \\
        \hline
        AskLosAngeles & 400 \\
        \hline
        AskNetsec & 995 \\
        \hline
        AskFeminists & 978 \\
        \hline
        AskWomenOver30 & 838 \\
       \hline
    \end{tabular}
    \caption{Number of questions we extracted from each SubReddit}
    \label{tab:subreddit-question-num}
\end{table}

\begin{table}[t]
    \centering
    \begin{tabular}{|c|c|c|}
    \hline
      ankle & arm & breast \\
      \hline
      cheeks & chin & collar bone\\
      \hline
      ear lobe & ear & elbow \\
      \hline
      eyebrows & eyelashes & eyelids\\
      \hline
      eyes & finger & foot \\
      \hline
      forehead & groin & hair \\
      \hline
      hand & heart & hip \\
      \hline
      intestines & jaw & knee \\
      \hline
      lips & liver & lungs \\
      \hline
      mouth & neck & nipple \\
      \hline
      nose & nostril & pancreas \\
      \hline
      pelvis & rectum & ribs \\
      \hline
      shin & shoulder blade & shoulder\\
      \hline
      spinal cord & spine & stomach \\
      \hline
      teeth & thigh & throat \\
      \hline
      thumb & toes & tongue \\
      \hline
      waist & wrist & \\
      \hline
    \end{tabular}
    \caption{Human Avatar Front Body Parts}
    \label{tab:front_bodyparts}
\end{table}

\begin{table}[t]
    \centering
    \begin{tabular}{|c|c|c|}
    \hline
      ankle & anus & arm \\
      \hline
      back & brain & buttocks \\
      \hline
      calf & ear lobe & ear \\
      \hline
      elbow & finger & foot \\
      \hline
      heart & intestines & kidney \\
      \hline
      knee & liver & lungs \\
      \hline
      neck & palm & pancreas \\
      \hline
      pelvis & rectum & ribs \\
      \hline
      scalp & shoulder blade & shoulder \\
      \hline
      spinal cord & spine & stomach \\
      \hline
      thigh & thumb & wrist \\
      \hline
    \end{tabular}
    \caption{Human Avatar Back Body Keywords. Some body parts can be visualized from both the front and back.}
    \label{tab:back_bodyparts}
\end{table}

\end{document}